\documentclass[sigconf]{acmart}

\usepackage[utf8]{inputenc} % allow utf-8 input
\usepackage[T1]{fontenc}    % use 8-bit T1 fonts
\usepackage{hyperref}       % hyperlinks
\usepackage{url}            % simple URL typesetting
\usepackage{booktabs}       % professional-quality tables
\usepackage{amsfonts}       % blackboard math symbols
\usepackage{nicefrac}       % compact symbols for 1/2, etc.
\usepackage{microtype}      % microtypography
\usepackage{xcolor}         % colors

\usepackage{multirow}
\usepackage[normalem]{ulem}
\usepackage{enumitem}
\usepackage{graphicx}
\usepackage{subcaption}

\usepackage{amsmath}

\usepackage{rotating}
\usepackage{multirow}
\usepackage{bbding}
\usepackage{makecell}

\usepackage{float}

\usepackage[ruled,vlined, linesnumbered]{algorithm2e}

\setlist[itemize]{nosep, left=10pt}

\AtBeginDocument{%
  }

\copyrightyear{2024}
\acmYear{2024}
\setcopyright{acmlicensed}
\acmConference[CIKM '24] {Proceedings of the 33rd ACM International Conference on Information and Knowledge Management}{October 21--25, 2024}{Boise, ID, USA.}
\acmBooktitle{Proceedings of the 33rd ACM International Conference on Information and Knowledge Management (CIKM '24), October 21--25, 2024, Boise, ID, USA}
\acmISBN{979-8-4007-0436-9/24/10}
\acmDOI{10.1145/3627673.3679568}

\begin{document}

%\title{Attention-DTDG: Transformer-Based Modeling for Learning Representation of Discrete-Time Dynamic Graphs}

\title{DTFormer: A Transformer-Based Method for Discrete-Time Dynamic Graph Representation Learning}

\author{Xi Chen}
\orcid{0009-0003-6156-4811}
\author{Yun Xiong}
\orcid{0000-0002-8575-5415}
\authornote{Corresponding author.}
\affiliation{%
  \institution{Shanghai Key Laboratory of Data Science, School of Computer Science, Fudan University}
  \city{Shanghai}
  \state{}
  \country{China}
}
\email{x\_chen21@m.fudan.edu.cn}
\email{yunx@fudan.edu.cn}

\author{Siwei Zhang}
\orcid{0009-0008-3994-083X}
\affiliation{%
  \institution{Shanghai Key Laboratory of Data Science, School of Computer Science, Fudan University}
  \city{Shanghai}
  \state{}
  \country{China}
}
\email{swzhang22@m.fudan.edu.cn}

\author{Jiawei Zhang}
\orcid{0000-0002-2111-7617}
\affiliation{%
  \institution{IFM Lab, Department of Computer Science, University of California, Davis}
  \city{}
  \state{CA}
  \country{USA}
}
\email{jiawei@ifmlab.org}

\author{Yao Zhang}
\orcid{0000-0003-1481-8826}
\affiliation{%
  \institution{Shanghai Key Laboratory of Data Science, School of Computer Science, Fudan University}
  \city{Shanghai}
  \state{}
  \country{China}
}
\email{yaozhang@fudan.edu.cn}

\author{Shiyang Zhou}
\orcid{0009-0001-6609-6020}
\author{Xixi Wu}
\orcid{0000-0002-9935-5957}
\affiliation{%
  \institution{Shanghai Key Laboratory of Data Science, School of Computer Science, Fudan University}
  \city{Shanghai}
  \state{}
  \country{China}
}
\email{22210240079@m.fudan.edu.cn}
\email{21210240043@m.fudan.edu.cn}

\author{Mingyang Zhang}
\orcid{0000-0001-6517-2880}
\author{Tengfei Liu}
\orcid{0000-0003-2871-4569}
\author{Weiqiang Wang}
\orcid{0000-0002-6159-619X}
\affiliation{%
  \institution{Ant Group}
  \city{Hangzhou}
  \state{}
  \country{China}
}
\email{zhangmingyang.zmy@antgroup.com}
\email{aaron.ltf@antgroup.com}
\email{weiqiang.wwq@antgroup.com}

\renewcommand{\shortauthors}{Xi Chen et al.}
\newcommand{\ours}{DTFormer}

\begin{abstract}
Discrete-Time Dynamic Graphs (DTDGs), which are prevalent in real-world implementations and notable for their ease of data acquisition, have garnered considerable attention from both academic researchers and industry practitioners. The representation learning of DTDGs has been extensively applied to model the dynamics of temporally changing entities and their evolving connections.
Currently, DTDG representation learning predominantly relies on GNN+RNN architectures, which manifest the inherent limitations of both Graph Neural Networks (GNNs) and Recurrent Neural Networks (RNNs). 
GNNs suffer from the over-smoothing issue as the models architecture goes deeper, while RNNs struggle to capture long-term dependencies effectively. GNN+RNN architectures also grapple with scaling to large graph sizes and long sequences.
Additionally, these methods often compute node representations separately and focus solely on individual node characteristics, thereby overlooking the behavior intersections between the two nodes whose link is being predicted, such as instances where the two nodes appear together in the same context or share common neighbors.

This paper introduces a novel representation learning method \ours~for DTDGs, pivoting from the traditional GNN+RNN framework to a Transformer-based architecture. Our approach exploits the attention mechanism to concurrently process topological information within the graph at each timestamp and temporal dynamics of graphs along the timestamps, circumventing the aforementioned fundamental weakness of both GNNs and RNNs. Moreover, we enhance the model's expressive capability by incorporating the intersection relationships among nodes and integrating a multi-patching module.
Extensive experiments conducted on six public dynamic graph benchmark datasets confirm our model's efficacy, achieving the SOTA performance.
\end{abstract}






\maketitle

\section{Introduction}
Dynamic graphs, characterized by their ability to the dynamics of temporally changing entities and their evolving interactions, find extensive applications across diverse fields including fraud detection \cite{khazane2019deeptrax}, recommendation systems \cite{you2019hierarchical}, and e-commerce \cite{yang2020dynamic}.
In dynamic graphs, entities are modeled as nodes, and their interactions are represented as edges, which typically encapsulate temporal information to accurately reflect interaction dynamics.
Recent advancements in dynamic graph methods \cite{tgcn, roland, tgat, tgn} focus on deriving the nodes' representations by aggregating their historical neighbor information. These nodes' representations are then processed via neural networks to predict future interactive behaviors.

Representation learning methods for dynamic graphs typically can be categorized into two distinct types, \textit{i.e.,} Discrete-Time Dynamic Graph (DTDG) and Continuous-Time Dynamic Graph (CTDG) \cite{barros2021survey, kazemi2020representation}, each employing unique approaches for graph modeling and learning node embeddings.
DTDGs \cite{tgcn, roland} model interactions within specified times or periods using static graph representations, capturing dynamic changes through successive graph snapshots.
Meanwhile, CTDGs \cite{tgat, tgn} consider time as a continuous variable and utilize timestamps on edges to maintain temporal attributes, thereby directly linking time with interaction dynamics.
In this paper, we will employ the DTDGs to model the dynamic graphs to be studied, which offers greater advantages over continuous-time based graph modeling approaches. According to prior work \cite{zhang2023dyted, skarding2021foundations, yang2021discrete, aggarwal2014evolutionary}, DTDGs can greatly simplify data processing and analysis by offering a manageable representation \cite{yang2021discrete}, and they support the use of a wide range of existing graph algorithms designed for static or coarser representations, which are easier to interpret and analyze \cite{aggarwal2014evolutionary}.

Previous methods for learning representations in DTDGs predominantly employ GNN+RNN architectures \cite{evolvegcn, tgcn, roland, gcrn, seo2018structured}, which address both topological and temporal dimensions of the data. Initially, within these existing methods \cite{evolvegcn, tgcn, roland}, static graph neural networks, such as GNNs \cite{gnn} and Graph Convolutional Networks (GCNs) \cite{gcn}, are used to process static snapshots of the graph to capture topological relationships. These networks function by aggregating information from neighboring nodes, thereby generating node embeddings for each graph snapshot.
Based on such learned representations by GNNs, these methods further employ RNNs, such as Long Short-Term Memory (LSTM) \cite{lstm} and Gated Recurrent Units (GRU) \cite{gru}, to integrate these embeddings over time, capturing the dynamic evolution of node features. This integration facilitates the prediction of future edge states, \textit{i.e.,} the existence or absence of connections between nodes. Although these methodologies adeptly capture both topological and temporal information for DTDGs, they also inherit significant limitations and challenges inherent to both GNNs and RNNs. 

One significant challenge in GNNs is over-smoothing \cite{chen2020measuring, yang2020revisiting}, a problem that occurs as GNNs integrate increasingly extensive neighborhood information through their layers. Initially, GNNs aggregate neighboring node features to update node representations. While additional layers theoretically enhance the richness of these representations by incorporating broader contextual information, they also lead to homogenization of features across nodes. As the number of layers increases, this homogenization intensifies, causing node features to converge and become indistinguishably similar, \textit{i.e.,} the \textit{over-smoothing} problem. This over-smoothing effect ultimately diminishes the distinctiveness and expressive power of node representations within the GNN architecture.

RNNs often struggle to capture node-wise long-term dependencies, especially in scenarios involving extended temporal sequences \cite{sherstinsky2020fundamentals, ilore}.
This challenge primarily arises from the propensity for vanishing or exploding gradients during the backpropagation process, which can severely impede learning stability and efficiency. Although enhanced versions of RNNs, such as LSTM and GRU, partially mitigate these issues, they still encounter significant difficulties when dealing with particularly long sequences \cite{sherstinsky2020fundamentals}.

Additionally, compared with static graphs, DTDGs usually have much large sizes, since they involve the attributes and connections temporally changing along the discrete timestamps. When employing a GNN+RNN architecture for DTDG representation learning, the increasing number of nodes and edges in DTDGs necessitates storing the entire graph in GPU memory. This requirement can lead to an excessive burden on GPU memory and a high risk of Out-Of-Memory (OOM) problems. Moreover, as the number of snapshots in a DTDG increases, RNNs are required to process excessively long sequences, further escalating the demand for computational resources. Consequently, most existing models are typically evaluated on relatively small networks which contain only a few hundred nodes or transaction graphs with a limited number of edges, and are limited to handling a few hundred snapshots.

% Furthermore, current methods typically model information for individual target nodes without considering the interactions between two nodes. However, in dynamic graphs, having interacted at the same moment or sharing common neighbors significantly influences their representations. For example, in social networks, if two individuals both follow another person or like the same tweet, their association is strengthened. Yet, current methods often overlook this aspect or struggle to model such behaviors effectively.

% Furthermore, existing approaches predominantly focus on modeling information for individual nodes, often neglecting the intersections between node pairs. However, in dynamic graphs, simultaneous interactions or shared neighbors play a crucial role in shaping node representations. For instance, in social networks like Twitter, the association between two individuals is reinforced if they both follow the same person or like the same tweet within the same time period. Yet, this aspect is often overlooked or inadequately modeled by current methodologies.

To addressing the limitations previously discussed, this paper proposes an innovative representation learning method for DTDG, termed \ours. This method adopts a novel perspective by moving away from the commonly used GNN+RNN architecture towards a Transformer-based architecture. Taking advantage of the attention mechanism, our method concurrently captures both temporal dynamics and node-specific information.
Initially, we collect all historical first-hop neighbors of both the two nodes, which are the nodes between which we aim to predict a future link. We then organize the features of these neighbors into sequences for Transformer processing.
This involves extracting features for neighbors or edges across five categories: node features, edge features, time (\textit{i.e.,} snapshot), frequency of appearance in a given snapshot, and joint appearance in a snapshot as neighbors.
Subsequently, we aggregate these sequences using the attention mechanism to derive the comprehensive representations of the nodes.

\begin{figure*}[!h]
\centering
\includegraphics[width=1\linewidth]{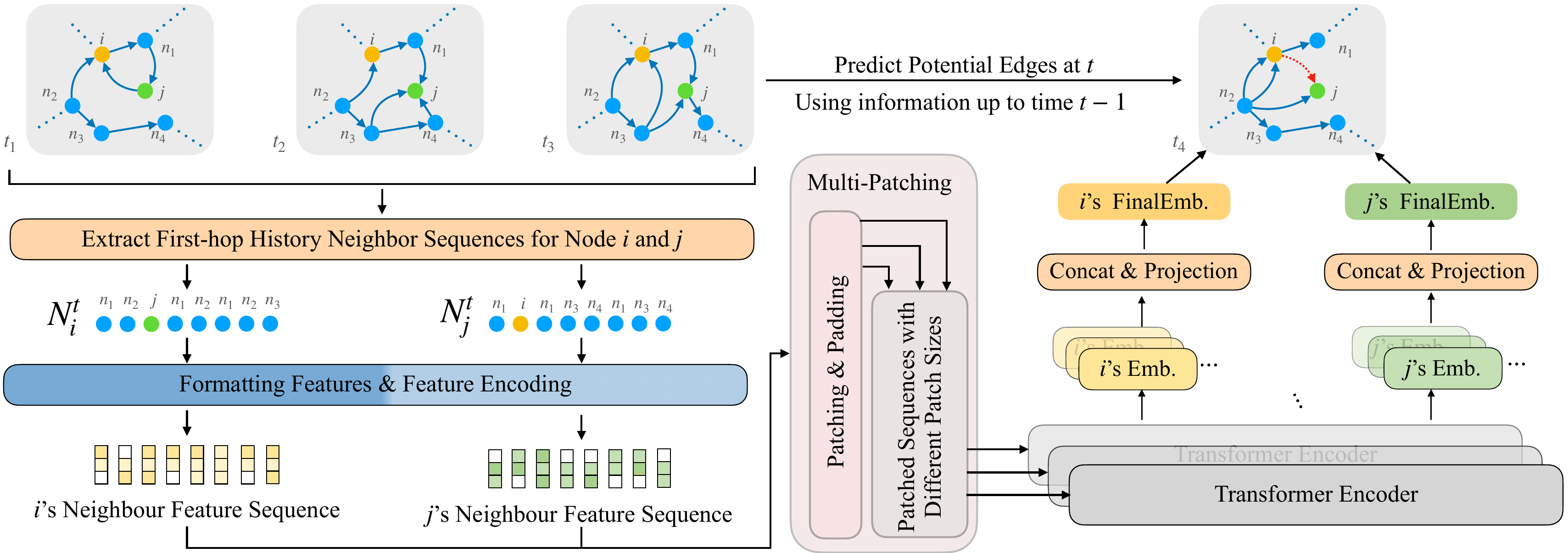}
\caption{Overview of \ours: First, we construct neighbor sequences for nodes $i$ and $j$ and format five distinct features, encoding them to create feature sequences. Next, we apply the multi-patching module and corresponding Transformer encoders to obtain the embeddings of nodes $i$ and $j$, which are then concatenated and used to predict the future link.}
\label{fig:model}
\end{figure*}

This method circumvents the problems prevalent in traditional GNN+RNN architectures, while enhancing the modeling of intersections between the two nodes.
The self-attention mechanism of the Transformer enables the model to focus more effectively on relevant nodes, helping to mitigate the over-smoothing issue. By adjusting the weights between nodes based on their relationships, the self-attention mechanism preserves the distinctiveness between nodes \cite{zhang2020graph}.
Furthermore, we introduce a novel multi-patching module that models dependencies at multiple temporal scales, effectively capturing both long-term and short-term patterns, while using far less memory space than previous methods.
Extensive experiments conducted across 6 public datasets demonstrate that our approach significantly surpasses previous state-of-the-art (SOTA) methods, as evidenced by improvements in the Mean Reciprocal Rank (MRR). Moreover, it facilitates more efficient training and inference processes, handling large-scale datasets characterized by numerous nodes, edges, and extensive temporal spans.

We summarize our contributions as follows:
\begin{itemize}
    \item We pioneer a departure from the conventional GNN+RNN framework for DTDG by treating interaction behaviors as sequence data, effectively circumventing the inherent limitations of traditional architectures.
    \item We present an innovative application of the Transformer to DTDG representation learning with our \ours~method, which enhances model expressiveness by incorporating five distinct feature types and utilizing a multi-patching module.
    \item By leveraging the advantages of the Transformer architecture and using our proposed multi-patching module to manage the length of input sequences, our method can handle larger datasets more efficiently while avoiding OOM issues.
    \item Our method is rigorously tested across 6 public datasets, with the results affirming its effectiveness and achieving unparalleled SOTA performance on these datasets.
\end{itemize}

%%%%%%%%%%%%%%%%%%%%%%%%%
\section{Related Work}
\label{sec:related_work}
\subsection{Dynamic Graphs Representation Learning}
\textbf{Discrete-Time Dynamic Graphs (DTDGs).}
Previous research on DTDGs predominantly employs combined architectures of GNNs and RNNs. These methodologies typically involve treating each temporal snapshot of the graph as a static graph, wherein neighbor node information is aggregated via GNNs, and RNNs are subsequently employed to capture temporal dynamics \cite{evolvegcn, tgcn, gcrn, seo2018structured}.
The latest model, ROLAND \cite{roland}, introduces a training method that aids the transition from static GNNs to dynamic models and features a live-update evaluation scenario where GNNs predict and update in a rolling manner.
It partially mitigates the OOM issue by reducing the amount of data stored on the GPU. However, this increases the burden of data transfer between GPU and CPU. Additionally, it is also based on the GNN+RNN architecture, which still suffers from other issues inherent to the GNNs and RNNs.

\noindent
\textbf{Continuous-Time Dynamic Graphs (CTDGs).}
% The study of CTDGs initiates with the Jodie \cite{jodie}, which utilizes dual RNNs to dynamically update node representations and employs a projection operator for estimating embeddings of nodes. This groundwork paves the way for TGN-based methods, exemplified by TGAT \cite{tgat}, which transforms CTDG models through the integration of an attention mechanism and substitutes the conventional positional encoding with time encoding, enhancing neighbor information aggregation. TGN \cite{tgn} consolidates previous technologies into a unified framework and proposes a memory module to archive nodes' historical interactions. Following these models, subsequent research tackles CTDG-specific challenges such as large-scale training \cite{tgl, speed}, noise dynamics \cite{rdgsl}, and node-wise long-term modeling \cite{dygformer, ilore}, addressing various complexities inherent in these methods.
The study of CTDGs began with Jodie \cite{jodie}, which uses dual RNNs and a projection operator to dynamically update node representations. This foundation led to TGN-based methods like TGAT \cite{tgat}, which integrates an attention mechanism to enhance neighbor information aggregation. TGN \cite{tgn} unifies previous technologies into a single framework and introduces a memory module to store nodes' historical interactions. 
%%%%%%%and replaces conventional positional encoding with time encoding to enhance neighbor information aggregation. TGN \cite{tgn} unifies previous technologies into a single framework and introduces a memory module to store nodes' historical interactions. 
Subsequent research mainly focuses on addressing CTDG-specific challenges \cite{tiger, tgl, speed, rdgsl, dygformer, ilore, chen2024prompt, zhang2024towards} such as large-scale training \cite{tgl, speed}, noise dynamics \cite{rdgsl}, and node-wise long-term modeling \cite{dygformer, ilore}.
% Subsequent research addresses CTDG-specific challenges such as large-scale training \cite{tgl, speed}, noise dynamics \cite{rdgsl}, and node-wise long-term modeling \cite{dygformer, ilore}.

\subsection{Transformers for Graph Learning}
The Transformer \cite{attention}, a groundbreaking model designed for sequential data analysis, leverages a self-attention mechanism to handle extended sequences. This capability is crucial for modeling long sequences effectively. The Transformer has already proven its versatility across various applications, including computer vision \cite{carion2020end, vit, liu2021swin}, natural language processing \cite{devlin2018bert, liu2019roberta, battaglia2018relational}, and time series analysis \cite{wu2021autoformer, zhou2021informer, li2019enhancing}. In the context of static graphs, numerous Transformer-based approaches have been developed \cite{rampavsek2022recipe, zhang2020graph, shi2022benchmarking, liu2022mask, dwivedi2020generalization, wang2022quest}. 
% One notable innovation is the introduction of a graph transformer layer that incorporates Laplacian Eigenvectors to represent graph structures \cite{dwivedi2020generalization}. Another significant advancement is the use of a graph transformer attention layer to distill information and understand neighborhood interactions, which demonstrates substantial efficacy \cite{wang2022quest}.

Despite the prevalence of GNN+RNN frameworks in DTDG research, applications of the Transformer as a core architecture remain relatively unexplored. Our model, however, integrates the Transformer into DTDG, pioneering new methodologies in DTDG modeling and potentially expanding the horizon of this field.

\section{Preliminaries}
\textbf{Discrete-Time Dynamic Graph as Sequences.}
Previous methods typically employ graph neural networks combined with sequence models (such as RNN-based models) to generate node embeddings for dynamic graphs. Due to the inherent weakness and limitations of both GNNs and RNNs as mentioned before, our approach discards such prior approaches and proposes a novel Transformer-based model in this paper instead.

For a series of graph snapshots of a DTDG, $\mathcal{G} = \{G_1, G_2, \ldots, G_T \}$, each individual graph snapshot can be considered as a static graph snapshot $G_t = (V_t, E_t)$, where $1 \leq t \leq T$. The node set of the graph snapshot $G_t$ is defined as $V_{t} = \{v_1, v_2, \ldots, v_n\}$ and the edge set as $E_t=\{e_{ij} = (i, j, t) \mid i, j \in V_{t}\}$. Nodes can be paired with features \( \mathbf{NODE}_t = \{\mathbf{node}_i \mid i \in V_t \} \), where feature vector $\mathbf{node}_i \in \mathbb{R}^{d_{N}}$ of node $i$ has dimension $d_{N}$. While edges can also have features \( \mathbf{EDGE}_t = \{\mathbf{edge}_{ij} \mid e_{ij} \in E_t\} \), where feature vector $\mathbf{edge}_{i,j} \in \mathbb{R}^{d_{E}}$ has dimension $d_{E}$.
To provide a sequence input for the Transformer-based encoder, we innovatively propose to use the history of first-hop neighbors of a node as the basic elements of the sequence. For a node $i$ in a snapshot $G_t$, we form a sequence by arranging all first-hop neighbors that have had an edge with node $i$ in any snapshot prior to $t$, ordered by the snapshot in which they appeared, denoted as $N_i^{t} = [n \mid \forall (i, n, {t'}) \text{ and } \forall t' < t ]$, where $(i, n, t')$ indicates an edge between $i$ and $n$ at snapshot $G_{t'}$. 
A specific neighbor $n$ can appear multiple times in $N_i^t$, because a single neighbor can have multiple edges with node $i$.
The sequence $N_i^t$ effectively captures the set of neighbors who has interacted with $i$ prior to the current time $t$, which effectively defines the learning contextual information for node $i$'s both within graphs and also across the time. Consequently, for all nodes, at different snapshots, we can obtain such a sequence, and the encoded information of this sequence through the Transformer-based encoder serves as the embedding of the node in the current snapshot.

\textbf{Problem Definition.}
Given a series of history snapshots 

\noindent
$\{ G_1, G_2, \ldots, G_{t-1} \}$ prior to time $t$, the goal is to predict whether there will be an edge between two nodes $i$ and $j$ in a future snapshot $G_t$ by utilizing the node embeddings of $i$ and $j$.

\section{Method}
% For predicting the potential edge between two nodes $i$ and $j$ in a snapshot at time $t$, we first characterize these nodes by embeddings through historical information and then use a projection head to predict the existence or absence for their future link. We construct two neighboring node sequences for the two nodes $i$ and $j$ from all snapshots prior to time $t$, sorting the first-hop neighbors by time and constructing sequences. These sequences are used to compute the node embeddings for $i$ and $j$. For instance, one sequence includes all first-hop neighbors of the node $i$. 
% We acquire the snapshot in which a neighboring node resides, utilizing it as a positional feature. Additionally, we calculate the number of occurrences of each neighbor node in all snapshots prior to time $t$, and use this data as the occurrence feature. We also consider the intersections between the neighbors of $i$ and $j$, forming an intersect feature. These elements, combined with the node and edge features from the source data, create an input sequence for the Transformer-based backbone model.
% This input sequence is then processed through a multi-patching method to create input patches, standardizing the input dimensions for all nodes. Subsequently, several Transformer-based encoders output node representations at various granularities. These representations are then integrated through a neural network to obtain the final node embeddings. The overall model is illustrated in Figure \ref{fig:model}. Next, we will provide detailed descriptions of each module in the model.

To predict a potential edge between two nodes $i$ and $j$ in a snapshot at time $t$, we characterize these nodes using embeddings derived from historical data. We start by extracting all first-hop historical neighbor nodes of \(i\) and \(j\), from which we construct five distinct feature sequences. A novel multi-patching module then segments and standardizes these sequences at different granularities. Patched sequences of various granularities are processed through Transformer-based backbone models, effectively capturing historical information and acquiring embeddings of nodes \(i\) and \(j\) at corresponding granularities. Using the node embeddings of \(i\) and \(j\), a projection head is then used to predict the presence or absence of their future link. The overall model structure is illustrated in Figure \ref{fig:model}. In this section, we first explain how we construct the feature sequences, followed by a description of the multi-patching module and the employed Transformer encoder. Finally, we discuss the implementation of the link prediction task.

\begin{figure}[!t]
\centering
\includegraphics[width=1\linewidth]{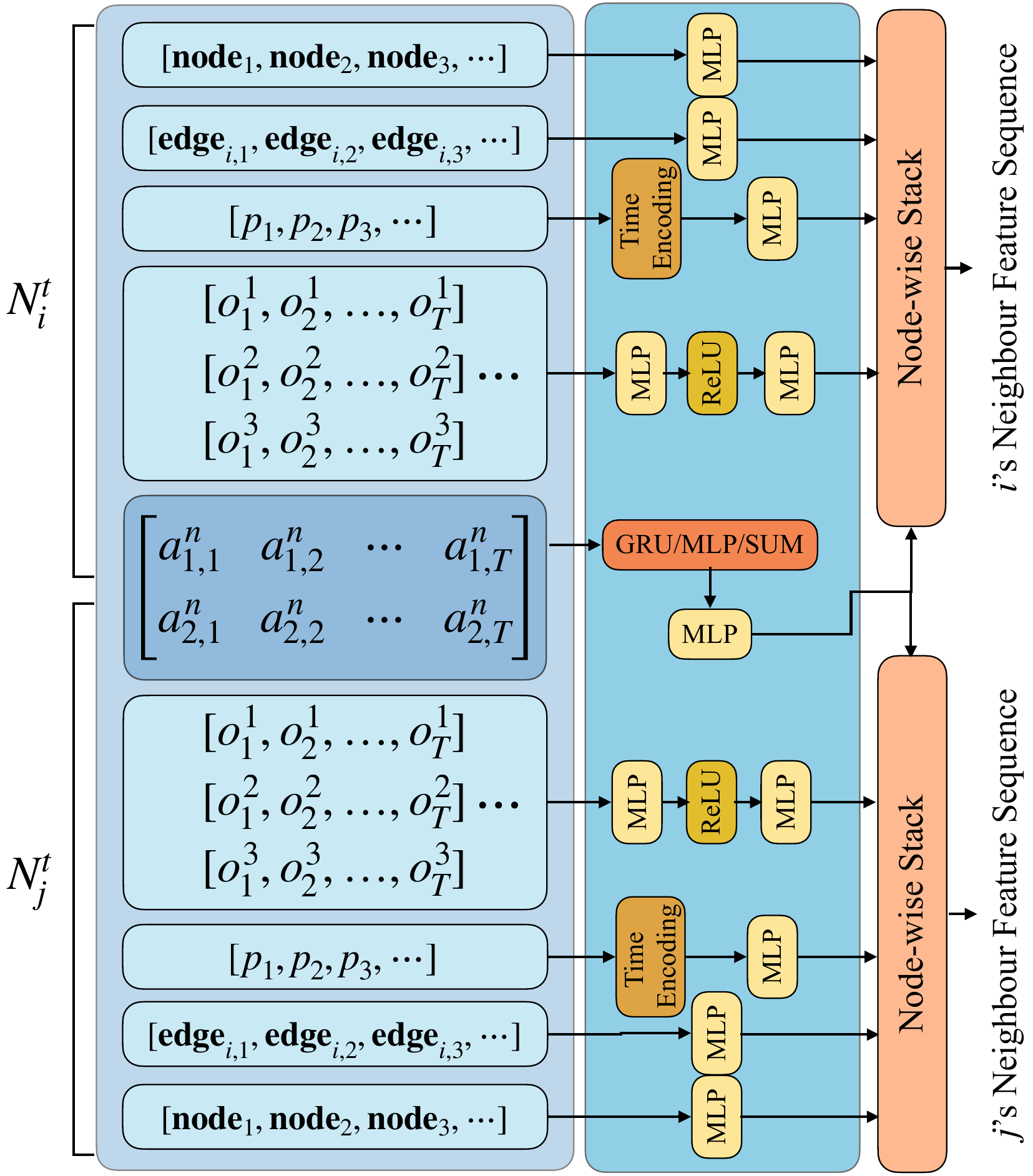}
\caption{Formatting and Encoding Features: For nodes $i$ and $j$, we format node, edge, positional, occurrence, and intersect features and encode them to create feature sequences.}
\label{fig:features}
\end{figure}

\subsection{Formatting Features}
An overview of formatting features is shown in Figure \ref{fig:features}.

\noindent
\textbf{Neighbor Basic Features.}
For a node $i$, the sequence of its first-hop neighbor nodes before time $t$,  $N_i^{t}$, can be obtained from the raw data, which includes the features of all neighboring nodes and their corresponding edge features. These are compiled into two corresponding sequences which lengths equal to the number of neighbor nodes $| N_i^{t} |$, denoted as: $\mathbf{NODE}_{N_i^{t}} = [\mathbf{node}_n]_{n \in N_i^{t}} \in \mathbb{R}^{| N_i^{t} | \times d_{N}}$ and $\mathbf{EDGE}_{N_i^{t}} = [\mathbf{edge}_{(i,n,t')}]_{n \in N_i^{t}} \in \mathbb{R}^{| N_i^{t} | \times d_{E}}$,
% \begin{equation*}
%     \mathbf{NODE}_{N_i^{t}} = [\mathbf{node}_n]_{n \in N_i^{t}} \in \mathbb{R}^{| N_i^{t} | \times d_{N}},
% \end{equation*}
% \begin{equation*}
%     \mathbf{EDGE}_{N_i^{t}} = [\mathbf{edge}_{(i,n,t')}]_{n \in N_i^{t}} \in \mathbb{R}^{| N_i^{t} | \times d_{E}},
% \end{equation*}
% and $\mathbf{EDGE}_{N_i^{t}} = [\mathbf{edge}_{(i,n,t')}]_{n \in N_i^{t}} \in \mathbb{R}^{| N_i^{t} | \times d_{E}}$, 
where $d_{N}$ and $d_{E}$ are the dimension of node features and edge features. These two sequences are then used as the basic features for the node $i$.

\noindent
\textbf{Neighbor Positional Feature.}
We assign sequential numbers to each graph snapshot according to their chronological order, using these positional indices in the order as substitutes for the traditional positional attributes in Transformers. The encoding of these attributes is then used as positional embedding inputs in the subsequent Transformer encoder. For a node $i$, every edge connecting it to its first-hop neighbors appears in an indexed graph snapshot. We obtain the index $p$ of the snapshot where an neighboring edge exists and apply an encoding function to this index.

There is an absence of explicit temporal attribute information in DTDG data. However, the chronological ordering of each snapshot inherently embeds temporal sequential information. This arrangement facilitates the incorporation of temporal order into our method. Inspired by the technique outlined in TGAT \cite{tgat}, we adopt a time encoding function similar to that described in TGAT to encode $p$, formulated as 
\begin{equation}
    \mathbf{P} = \sqrt{\frac{1}{2 \cdot d_{P}}} \left[\cos(w_1 p), \sin(w_1 p), \ldots, \cos(w_{d_{P}} p), \sin(w_{d_{P}} p)\right], 
\end{equation}
which serves as the positional feature for the neighboring nodes sequence. Here the dimension of the positional feature $\mathbf{P}$ will be $2 \cdot d_{P}$ and $w_1,\cdots,w_{d_P}$ are trainable parameters. This approach effectively leverages temporal information, thereby enhancing the encoder’s ability to process and interpret time-related data attributes. Similarly, this positional feature can also be organized into a sequence of length equal to $| N_i^{t} |$, denoted as: $\mathbf{POS}_{N_i^{t}} = [\mathbf{P}_n]_{n \in N_i^{t}} \in \mathbb{R}^{| N_i^{t} | \times 2 \cdot d_{P}}.$
% $\mathbf{POS}_{N_i^{t}} = \{\mathbf{P}_1, \mathbf{P}_2, \ldots, \mathbf{P}_{| N_i^{t} |} \} \in \mathbb{R}^{| N_i^{t} | \times 2 \cdot d_{P}}$. 
% \begin{equation*}
%     \mathbf{POS}_{N_i^{t}} = [\mathbf{P}_n]_{n \in N_i^{t}} \in \mathbb{R}^{| N_i^{t} | \times 2 \cdot d_{P}}.
% \end{equation*}
Using time encoding to encode the snapshot index enables us to utilize the limited temporal-related information in DTDGs, thereby enhancing the model’s capacity.

\noindent
\textbf{Neighbor Occurrence Feature.}
To enhance the representation of nodes and capture the frequency and pattern of interactions of neighbors in their historical behavior, we design an occurrence feature. For each first-hop neighbor of a node $i$, we compute the number of their interactions in each snapshot, resulting in an occurrence vector \( \mathbf{O}_n = [o_1^n, o_2^n, \dots, o_T^n] \in \mathbb{R}^{T} \) of length $T$, where $T$ is the  total number of snapshots. Here \( o_t^n \) represents the number of interactions between node \( i \) and its neighbor \( n \) at snapshot \( t \), for \( t = 1, 2, \dots, T \). Subsequently, in accordance with the snapshot currently being predicted, we mask the occurrence vector to retain only the features prior to this snapshot to prevent information leakage. The masked occurrence vector is noted as $\bar{\mathbf{O}}_{n}$, and then used as the occurrence feature. Similarly, the occurrence features of all neighbor nodes can form a sequence, denoted as: $\mathbf{OCC}_{N_i^{t}} = [\bar{\mathbf{O}}_{n}]_{n \in N_i^{t}} \in \mathbb{R}^{| N_i^{t} | \times T}.$
% \begin{equation*}
%     \mathbf{OCC}_{N_i^{t}} = [\bar{\mathbf{O}}_{n}]_{n \in N_i^{t}} \in \mathbb{R}^{| N_i^{t} | \times T}.
% \end{equation*}
% \mathbf{OCC}_{N_i^{t}} = [\bar{\mathbf{O}}_{n}]_{n \in N_i^{t}} \in \mathbb{R}^{| N_i^{t} | \times T}$.

\noindent
\textbf{Neighbor Intersect Feature.}
\label{sec:ints_feat}
Existing approaches predominantly focus on modeling information for individual nodes, often neglecting the intersections between node pairs. However, in dynamic graphs, simultaneous interactions or shared neighbors play a crucial role in shaping node representations. Intuitively, if two nodes have shared common neighbors at some historical moment, the likelihood of new interaction between these two nodes in the future may increase with the interaction frequency of these common neighbors. For instance, in social networks like Twitter, the association between two individuals is reinforced if they both follow the same person or like the same tweet within the same time period. Furthermore, this regularity in the data may exhibit periodicity, For instance, in an e-commerce networks, individuals who share an interest in camping often tend to purchase related items predominantly during weekends. This observation suggests a cyclical pattern in consumer behaviors that could be crucial for predicting future purchase trends.

Yet, this aspect is often overlooked or inadequately modeled by current methodologies.
To model this scenario and incorporate the temporal sequence information of dynamic graphs, we constructed an intersect feature for nodes $i$ and $j$. Initially, for one of the neighbor nodes $n$, we build a matrix $\mathbf{A}_n$ of dimensions $\mathbb{R}^{2 \times T}$, where the first row stores the count of occurrences of the current neighbor node among the neighbors of node $i$, categorized by snapshot to indicate in which snapshots this neighbor node has appeared. The second row stores the count of occurrences of the current neighbor node as a neighbor of the other node $j$, also categorized by snapshot. The matrix $\mathbf{A}_n$ is defined as follows:
\[
\mathbf{A}_n = \begin{bmatrix}
a_{1,1}^n & a_{1,2}^n & \cdots & a_{1,T}^n \\
a_{2,1}^n & a_{2,2}^n & \cdots & a_{2,T}^n
\end{bmatrix},
\]
where: \( a_{1,t}^n \) represents the number of times neighbor \( n \) appears as a neighbor of node \( i \) in snapshot \( t \), and \( a_{2,t}^n \) represents the number of times neighbor \( n \) appears as a neighbor of node \( j \) in snapshot \( t \), for \( t = 1, 2, \dots, T \).
This matrix thus records the counts of the current neighbor node as a neighbor for both nodes $i$ and $j$, separated according to the order of the snapshots.
%For example, node $i$ has a neighbor sequence $a,b,c,c$, the corresponding snapshot indexes for them are $1,2,2,3 $, and node $j$ has a neighbor sequence $b,a,c,a$ at $2,3,3,3$, the  $\mathbf{I}_i$ should be $[[[1,0],[0,0],[0,2]], [[0,0],[1,1],[0,0]], [[0,0],[1,0],[1,1]] ]$.
All $\mathbf{A}_n$ matrices for the neighbors are then processed through a neural network (sequence networks like GRU or simply MLP), denoted as $f(\cdot)$, to represent the features $ \bar{\mathbf{A}} = f(\mathbf{A})$ and normalize them to a dimension $\mathbb{R}^{d_{I}}$, where $d_{I}$ is the dimension of the intersect feature. We explore three different modes of $f(\cdot)$: GRU mode, MLP mode, and SUM mode. This will be detailed discussed in the ablation study section, Section \ref{sec:different_sif}. Subsequently, for all neighbors of a node, their $\bar{\mathbf{A}}$ can form a sequence, serving as the intersect feature for that node, denoted as: $\mathbf{INT}_{N_i^{t}} = [\bar{\mathbf{A}}_n]_{n \in N_i^{t}} \in \mathbb{R}^{|N_i^{t}| \times d_{I}}.$
% \begin{equation*}
%     \mathbf{INT}_{N_i^{t}} = [\bar{\mathbf{A}}_n]_{n \in N_i^{t}} \in \mathbb{R}^{|N_i^{t}| \times d_{I}}.
% \end{equation*}
% $\mathbf{INT}_{N_i^{t}} = [\bar{\mathbf{A}}_n]_{n \in N_i^{t}} \in \mathbb{R}^{|N_i^{t}| \times d_{I}}$.

\begin{figure}[!t]
\centering
\includegraphics[width=1\linewidth]{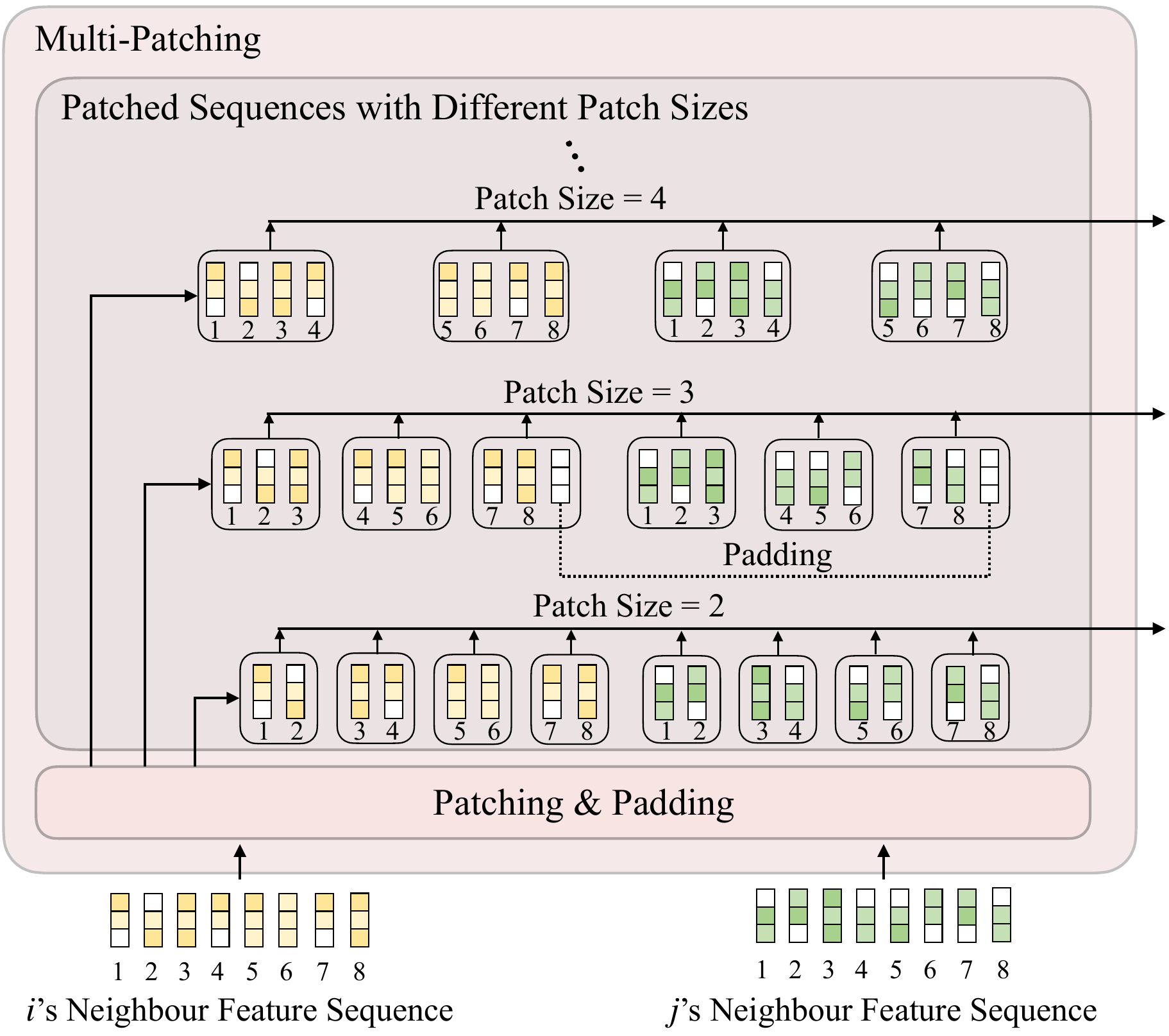}
\caption{Multi-Patching: Applying multiple patch sizes to split feature sequences helps our model capture information at different granularities and reduce model complexity.}
\label{fig:multi_patching}
\end{figure}

\subsection{Multi-Patching Module}
To reduce the computational complexity of the subsequent Transformer encoder, we introduce the patching operation first proposed in \cite{vit} into our model. This operation helps Transformer encoder captures local information in sequence data. Typically, smaller patches might focus on capturing fine-grained local features, while larger patches could encompass more contextual information \cite{dygformer}. To simultaneously capture information at different granularities, corresponding to different lengths of time segments in dynamic graphs, we further design a multi-patching module.

This module involves segmenting the existing sequences into patches of various sizes to obtain multiple subsequences. Let $S_m$ denote the patch size, and $\lambda_m$ denote the length of new sequence, each new sequence is of length equal to the original sequence length divided by the patch size: $\lambda_m = | N_i^{t} | / S_m$. If \(| N_i^{t} |\) cannot be divided by \(S_m\), padding will be applied to the end of the patched sequence. Given different patch sizes $S_m, S_n, \ldots$, the lengths of the new sequences vary $\lambda_m, \lambda_n, \ldots$. For a node $i$, assume the patch size is $S_m$, each of feature sequences undergoes the patching operation, we now have five corresponding new patched sequences: $\mathbf{NODE}_{N_i^{t}} \in \mathbb{R}^{\lambda_i^m \times d_{N}}$, $\mathbf{EDGE}_{N_i^{t}} \in \mathbb{R}^{\lambda_i^m \times d_{E}}$, $\mathbf{POS}_{N_i^{t}} \in \mathbb{R}^{\lambda_i^m \times 2 \cdot d_{P}}$,
$\mathbf{OCC}_{N_i^{t}} \in \mathbb{R}^{\lambda_i^m \times T}$ and
$\mathbf{INT}_{N_i^{t}} \in \mathbb{R}^{\lambda_i^m \times d_{I}}$. 
For different patch sizes, resulting in multiple new patched sequences of varying granularity. We then use these multi-patched sequences as inputs for the next step of the Transformer encoders. An illustration of multi-patching module is shown in Figure \ref{fig:multi_patching}.

Due to the patching operation, the length of the input sequences to the subsequent Transformer encoder can be controlled, allowing us to maintain our model complexity within an appropriate range. The related complexity analysis is discussed in Section \ref{sec:exp_complx}. Additionally, the introduction of multi-patching enables our model to capture multi-granularity historical information simultaneously, thereby enhancing the expressive power of our model.

\subsection{Transformer Encoder}
In order to correspond with the multiple different patch sizes obtained through multi-patching, we employ multiple Transformer encoders, each designed to accept patched sequences of different lengths, thereby capturing information at varying granularities. The output of each Transformer encoder is considered as the representation of the node at the current granularity. Finally, these representations are integrated through a neural network (usually an MLP) to obtain the final embedding for the node.

Taking patched sequences at a specific granularity as an example, for nodes $i$, we first pass the five patched sequences $\mathbf{NODE}_{N_i^{t}}$, $\mathbf{EDGE}_{N_i^{t}}$, $\mathbf{POS}_{N_i^{t}}$,
$\mathbf{OCC}_{N_i^{t}}$ and $\mathbf{INT}_{N_i^{t}}$ through MLPs to standardize their dimensions, resulting in five uniformly dimensioned sequences:
% \begin{equation}
%     \mathbf{NODE}_{N_i^{t}}^C = \text{MLP}(\mathbf{NODE}_{N_i^{t}}) \in \mathbb{R}^{\lambda_m \times d_C},
% \end{equation}
% \begin{equation}
%     \mathbf{EDGE}_{N_i^{t}}^C = \text{MLP}(\mathbf{EDGE}_{N_i^{t}}) \in \mathbb{R}^{\lambda_m \times d_C},
% \end{equation}
% \begin{equation}
%     \mathbf{POS}_{N_i^{t}}^C = \text{MLP}(\mathbf{POS}_{N_i^{t}}) \in \mathbb{R}^{\lambda_m \times d_C},
% \end{equation}
% \begin{equation}
%     \mathbf{OCC}_{N_i^{t}}^C = \text{MLP} ( \text{ReLU}(\text{MLP}(\mathbf{OCC}_{N_i^{t}}))) \in \mathbb{R}^{\lambda_m \times d_C},
% \end{equation}
% \begin{equation}
%     \mathbf{INT}_{N_i^{t}}^C = \text{MLP}(\mathbf{INT}_{N_i^{t}}) \in \mathbb{R}^{\lambda_m \times d_C},
% \end{equation}
\begin{equation}
    \mathbf{FEAT*}_{N_i^{t}}^C = \text{MLP}(\mathbf{FEAT*}_{N_i^{t}}) \in \mathbb{R}^{\lambda_i^m \times d_C},
\end{equation}
\begin{equation}
    \mathbf{OCC}_{N_i^{t}}^C = \text{MLP} ( \text{ReLU}(\text{MLP}(\mathbf{OCC}_{N_i^{t}}))) \in \mathbb{R}^{\lambda_i^m \times d_C},
\end{equation}
where all these patched sequences will have the identical dimension $d_C$, and $\mathbf{Feat*}$ can be $\mathbf{NODE}$, $\mathbf{EDGE}$, $\mathbf{POS}$ or $\mathbf{INT}$.
Due to there is an extra dimension for occurance feature, we apply an extra MLP and ReLU to unify the dimension.
The five dimensionally uniform sequences are then concatenated to form the input sequence for the Transformer encoder: $\mathbf{C}_i^{\lambda_i^m} = \mathbf{NODE}_{N_i^{t}}^C \parallel \mathbf{EDGE}_{N_i^{t}}^C \parallel \mathbf{POS}_{N_i^{t}}^C \parallel\mathbf{OCC}_{N_i^{t}}^C \parallel \mathbf{INT}_{N_i^{t}}^C \in \mathbb{R}^{\lambda_i^m \times 5\cdot d_C}$, $\parallel$ indicates the concatenation operation.
This process is completed for both nodes, thus we have $\mathbf{C}_i^{\lambda_i^m}$ and $\mathbf{C}_j^{\lambda_j^m}$.

To capture the interactive dependencies between the two nodes, their sequences are stacked and fed into the Transformer \cite{attention, vit} for encoding:
\begin{equation}
    \mathbf{X}^{(0)} = [\mathbf{C}_i^{\lambda_i^m}, \mathbf{C}_j^{\lambda_j^m}].
\end{equation}
The Transformer encoder comprises $L$ layers, and the output of the Transformer encoder at the final layer $L$ is denoted by $\mathbf{H} = \mathbf{X}^{(L)}$. 
The entire process of the Transformer encoder can be summarized as follows:
\begin{equation}
    \mathbf{H}_{\lambda_m} = \text{Transformer}_{\lambda_m}([\mathbf{C}_i^{\lambda_i^m}, \mathbf{C}_j^{\lambda_j^m}]).
\end{equation}

\subsection{Link Prediction and Loss Function}

For different patch sizes and sequence lengths, the output of the Transformer encoder then serves as the representation of the two nodes. 
Since features of two different nodes are input simultaneously, after the output, based on the order of the output sequence, it is split into two sequences representing the respective nodes, and by averaging, the representations of the two nodes at the current granularity can be obtained:
\begin{equation}
    \mathbf{Emb}_i^{\lambda_i^m}, \mathbf{Emb}_j^{\lambda_j^m} = \text{MEAN}(\mathbf{H}_{\lambda_m}[:\lambda_i^m]), \text{MEAN}(\mathbf{H}_{\lambda_m}[\lambda_i^m:]).
\end{equation}

Finally, by concatenating the representations of the nodes at different granularities and passing them through an MLP, the final embeddings of the two nodes are obtained:
\begin{equation}
    \mathbf{Final Emb}_i = \text{MLP}(\mathbf{Emb}_i^{\lambda_i^m} \parallel \mathbf{Emb}_i^{\lambda_i^n} \parallel \cdots),
\end{equation}
\begin{equation}
    \mathbf{Final Emb}_j = \text{MLP}(\mathbf{Emb}_j^{\lambda_j^m} \parallel \mathbf{Emb}_j^{\lambda_j^n} \parallel \cdots).
\end{equation}

% \subsection{Link Prediction and Loss Function}

After acquiring the final embeddings of the nodes $i$ and $j$, a projection head, \textit{i.e.,} a two-layer MLP, is used to compute the probability of an edge existing between them, facilitating predictions for downstream tasks:
\begin{equation}
    \hat{Y}_{ij} = \text{MLP}(\mathbf{Final Emb}_i \parallel \mathbf{Final Emb}_j),
\end{equation}
and we use the cross-entropy as the loss function:
\begin{equation}
    \mathcal{L} = \text{Cross-Entropy}(\hat{Y}_{ij}, Y_{ij}).
\end{equation}

\useunder{\uline}{\ul}{}

\begingroup
\renewcommand{\arraystretch}{0.9}
\begin{table*}[h]
\centering

\caption{A performance comparison with baseline DTDG models on the future link prediction task is conducted, with MMRs reported. All experiments are conducted using three random seeds, and the average results and standard deviations are reported. Baseline results are derived from the ROLAND paper \cite{roland} and its corresponding implementation. We further report the AUC-ROC and AP metrics for our method to facilitate future comparisons by researchers.}

\resizebox{\textwidth}{!}{
\begin{tabular}{c|cccccc}
\toprule
Link Prediction   & Bitcoin-OTC           & UCI-Message           & Reddit-Title        & Reddit-Body         & Mathoverstack       & Email         \\ \midrule
GCN               & 0.0025                & 0.1141                & N/A, OOM            & N/A, OOM            & N/A, OOM            & N/A, OOM              \\
DynGEM            & 0.0921                & 0.1055                & N/A, OOM            & N/A, OOM            & N/A, OOM            & N/A, OOM              \\
dyngraph2vecAE    & 0.0916                & 0.0540                & N/A, OOM            & N/A, OOM            & N/A, OOM            & N/A, OOM              \\
dyngraph2vecAERNN & 0.1268                & 0.0713                & N/A, OOM            & OOM                 & N/A, OOM            & N/A, OOM              \\
EvolveGCN-H       & 0.067 ± 0.035         & 0.061 ± 0.040         & N/A, OOM            & 0.148 ± 0.013       & N/A, OOM            & 0.025 ± 0.016         \\
EvolveGCN-O       & 0.085 ± 0.022         & 0.071 ± 0.009         & N/A, OOM            & N/A, OOM            & N/A, OOM            & N/A, OOM              \\
GCRN-GRU          & OOM                   & 0.080 ± 0.012         & N/A, OOM            & N/A, OOM            & N/A, OOM            & 0.059 ± 0.052         \\
GCRN-LSTM         & OOM                   & 0.083 ± 0.001         & N/A, OOM            & N/A, OOM            & N/A, OOM            & 0.038 ± 0.010         \\
GCRN-Baseline     & 0.152 ± 0.011         & 0.069 ± 0.004         & N/A, OOM            & N/A, OOM            & N/A, OOM            & 0.074 ± 0.037         \\
TGCN              & 0.128 ± 0.049         & 0.054 ± 0.024         & N/A, OOM            & N/A, OOM            & N/A, OOM            & 0.058 ± 0.040       \\
ROLAND-Fix-GRU    & {\ul 0.2203 ± 0.0167} & {\ul 0.2289 ± 0.0618} & N/A, OOM            & N/A, OOM            & N/A, OOM            & 0.0207 ± 0.0158       \\
ROLAND-GRU-Update & 0.194 ± 0.004         & 0.112 ± 0.008         & {\ul 0.425 ± 0.015} & {\ul 0.362 ± 0.002} & {\ul 0.271 ± 0.045} & {\ul 0.102 ± 0.026} \\ \midrule
\ours~ (MRR)   & \textbf{0.2838 ± 0.0291}   & \textbf{0.4925 ± 0.0208}   & \textbf{0.4296 ± 0.0114}   &  \textbf{0.5055 ± 0.0166} &    \textbf{0.5431 ± 0.0328}  &  \textbf{0.2398 ± 0.0226} \\\midrule
Improvement over best baseline & {\color[HTML]{3166FF} \textbf{28.82\%}}   & {\color[HTML]{3166FF} \textbf{81.73\%}}   &  {\color[HTML]{3166FF} \textbf{1.08\%}}     &  {\color[HTML]{3166FF} \textbf{21.54\%}} &  {\color[HTML]{3166FF} \textbf{100.41\%}}     & {\color[HTML]{3166FF} \textbf{134.18\%}}       \\ \midrule
\ours~ (AUC-ROC)    & 0.9477 ± 0.0022    & 0.9524 ± 0.0024   &  0.9764 ± 0.0001    &  0.9746 ± 0.0002 & 0.9745 ± 0.0018     &   0.9670 ± 0.0006     \\
\ours~ (AP)    & 0.9527 ± 0.0008   & 0.9607 ± 0.0026   &  0.9792 ± 0.0001    & 0.9781 ± 0.0003 &  0.9780 ± 0.0014    &  0.9633 ± 0.0012      \\\bottomrule

\end{tabular}

}

\label{tab:baselines}
\end{table*}

\endgroup

\section{Experiments}
\subsection{Experimental Setup}

\textbf{Datasets.} We conduct extensive experiments on 6 public datasets: Bitcoin-OTC \cite{bitcoin}, UCI-Message \cite{collegemsg}, Reddit-Title \cite{reddit}, Reddit-Body \cite{reddit}, Mathoverstack \cite{emailmath}, and Email-Eu-core (Email) \cite{emailmath}. % Key metadata related to above datasets are summarized in Table \ref{tab:datasets}.

% \begin{itemize}
%     \item Bitcoin-OTC: This dataset includes who-trusts-whom networks of individuals bitcoin trading on the OTC platform.
%     \item UCI-Message: Comprising private messages among students within an online social network, this dataset captures interpersonal communications in an academic setting.
%     \item Reddit-Title and Reddit-Body: These datasets consist of networks formed by hyperlinks in the titles and bodies of Reddit posts, respectively, with each hyperlink representing a directed edge between two subreddits.
%     \item Mathoverstack: Derived from a question-and-answer platform, this dataset represents interactions through comments on posts or answers to specific questions.
%     \item Email: Originating from a European research institution, this dataset reflects the email communication patterns among its members.
% \end{itemize}

\noindent
\textbf{Baselines.} 
We employ various methods for DTDG representation learning as baselines, as detailed in Section \ref{sec:related_work}. The majority of these methods utilize the GNN+RNN model architecture. Specifically for ROLAND \cite{roland}, which incorporates multiple training approaches and diverse backbone models, we selected the two most effective setups as strong baselines.

\noindent
\textbf{Task.}
We adopt the future link prediction task, a standard evaluation used in previous studies, to evaluate our model. Specifically, we utilize historical data, \textit{i.e.,} snapshots up to time $t-1$, to predict the presence of edges at a subsequent snapshot at time $t$. Consistent with the methodology employed in the ROLAND model, we evaluate our models using the Mean Reciprocal Rank (MRR). For each node $i$ with a positive edge to node $j$ at time $t$, we randomly sample 1,000 negative edges originating from $i$, and ranked the positive edge $(i,j,t)$ based on its prediction score relative to these negative edges. The overall MRR score is calculated as the mean of the reciprocal ranks for all testing nodes $i$. Additionally, to enrich the comparative data for future research, we also report the Area Under the Receiver Operating Characteristic Curve (AUC-ROC) and the Average Precision (AP) for our method.

\noindent
\textbf{Implementation Details.}
For all baseline models, we adhered to the configurations and parameters specified in ROLAND. For models not covered in ROLAND, we utilize the settings outlined in their respective original papers. Our model employ a consistent data division of 80\% for training, 10\% for validation, and 10\% for testing. Training is set for 100 epochs, with a patience of 20 epochs for early stopping. We conduct all experiments using the Adam optimizer and repeat each experiment three times to ensure robust error estimation. 
Other hyperparameters may be adjusted according to different datasets to obtain optimized results. For example, with large datasets, a longer maximum input sequence length for the transformer may help the model gather more information about nodes from their neighbors. However, to simplify the experimental process, we use a unified set of hyperparameters. In our main experiments involving the multi-patching module, we employ three different patch sizes (2, 4, and 8) and set a maximum sequence length of 32 for all datasets.
All experiments are performed on a single server equipped with a 72-core CPU, 128GB of RAM, and 4 Nvidia Tesla V100 GPUs, each with 32GB of memory. The codes of our methods are available at \href{https://github.com/chenxi1228/DTFormer}{\uline{our GitHub repository}}.

\begin{figure*}[!tbp]
    \centering
    \begin{subfigure}[b]{.3\linewidth}
        \includegraphics[width=\linewidth]{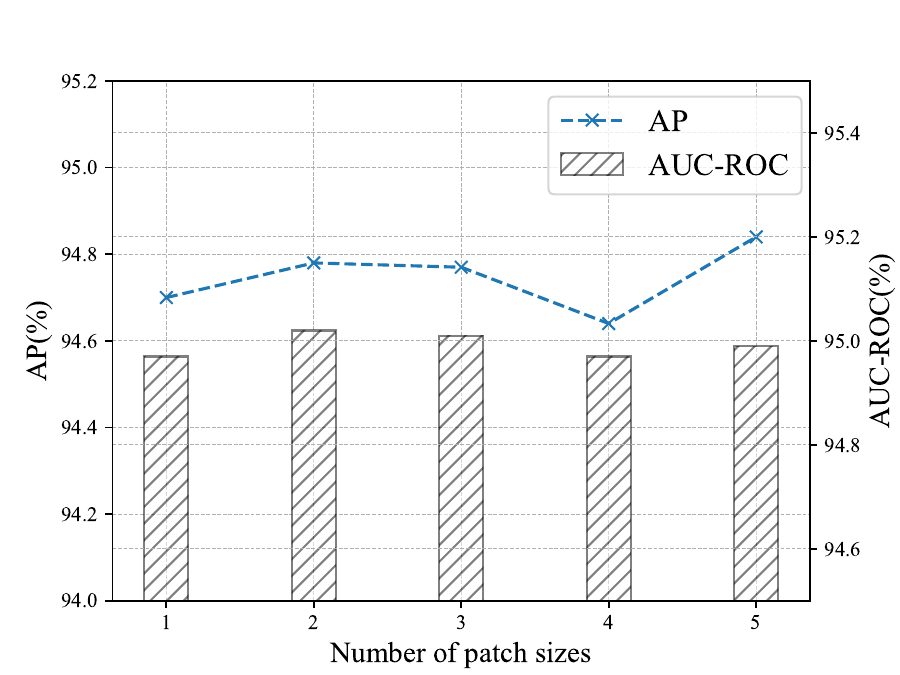}
        \caption{Bitcoin-OTC}
    \end{subfigure}
    \hspace{5pt}
    \begin{subfigure}[b]{.3\linewidth}
        \includegraphics[width=\linewidth]{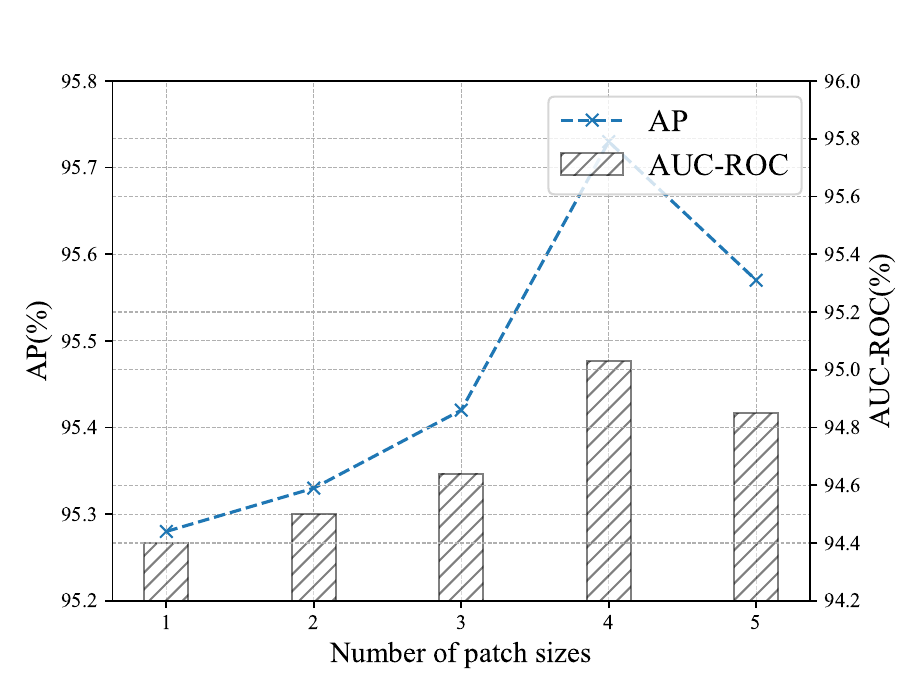}
        \caption{UCI-Message}
    \end{subfigure}
        \hspace{5pt}
    \begin{subfigure}[b]{.3\linewidth}
        \includegraphics[width=\linewidth]{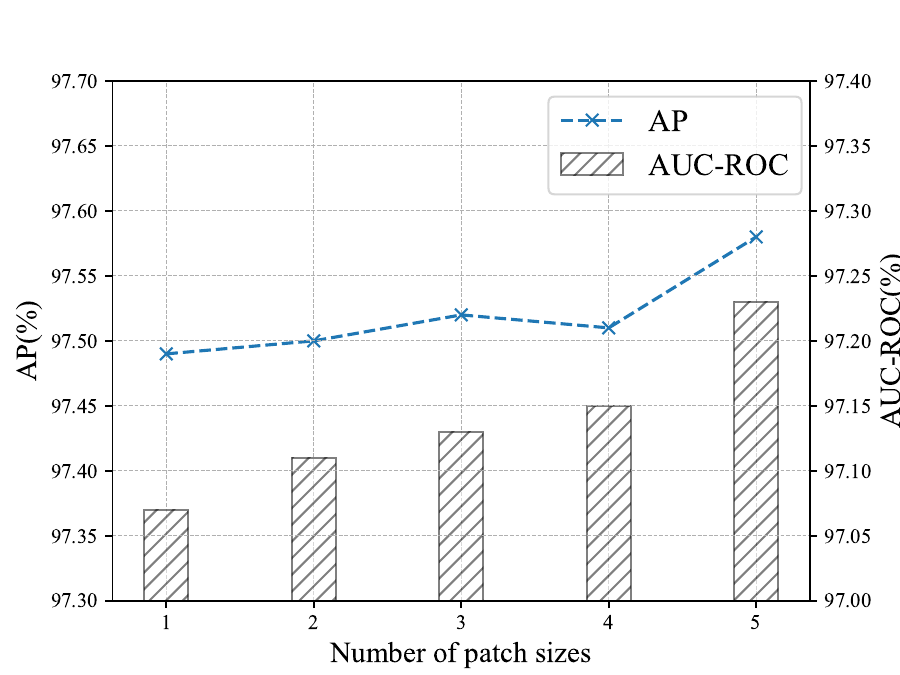}
        \caption{Reddit-Body}
    \end{subfigure}

\caption{Experiment results for the empirical study of the multi-patching module, reporting the AP(\%) and AUC-ROC(\%).}
\label{fig:mp}
\end{figure*}
\begin{table*}[]

\centering

\caption{Ablation Study: We report the performance of our method, measured by AUC-ROC and AP, when applying different neighbor features. In this experiment, we use the SUM mode to model the Neighbor Intersect Feature.}

\resizebox{\textwidth}{!}{
\begin{tabular}{c|cccc|cccccc}
\toprule
\multirow{2}{*}{Metric}                            & \multicolumn{4}{c|}{Features}                                         & \multirow{2}{*}{Bitcoin-OTC} & \multirow{2}{*}{UCI-Message} & \multirow{2}{*}{Reddit-Title} & \multirow{2}{*}{Reddit-Body} & \multirow{2}{*}{Mathoverstack} & \multirow{2}{*}{Email} \\
                                                   & Node\&Edge      & Pos.     & Occ.     & Int. &                              &                              &                               &                              &                                &                                \\ \midrule
\multirow{5}{*}{\rotatebox[origin=c]{90}{AUC-ROC}} & \CheckmarkBold & \XSolidBrush   & \XSolidBrush   & \XSolidBrush       & 0.5000 ± 0.0000              & 0.5000 ± 0.0000              & 0.9096 ± 0.0020               & 0.8920 ± 0.0015              & 0.5000 ± 0.0000                & 0.5000 ± 0.0000                \\
                                                   & \CheckmarkBold & \CheckmarkBold & \XSolidBrush   & \XSolidBrush       & 0.7394 ± 0.0056              & 0.7977 ± 0.0021              & 0.9475 ± 0.0002               & 0.9320 ± 0.0004              & 0.7895 ± 0.0006                & 0.8443 ± 0.0027                \\
                                                   & \CheckmarkBold & \XSolidBrush   & \CheckmarkBold & \XSolidBrush       & 0.8513 ± 0.0031              & 0.8476 ± 0.0072              & 0.9491 ± 0.0013               & 0.9358 ± 0.0013              & 0.8864 ± 0.0111                & 0.9565 ± 0.0009                \\
                                                   & \CheckmarkBold & \XSolidBrush   & \XSolidBrush   & \CheckmarkBold     & 0.9075 ± 0.0017              & 0.9446 ± 0.0004              & 0.9634 ± 0.0004               & 0.9656 ± 0.0001              & 0.9204 ± 0.0027                & 0.9817 ± 0.0002                \\ 
                                                    & \CheckmarkBold & \CheckmarkBold   & \CheckmarkBold   & \CheckmarkBold     & 0.9477 ± 0.0022 & 0.9524 ± 0.0024 & 0.9764 ± 0.0001   & 0.9746 ± 0.0002  &    0.9350 ± 0.0023  &     0.9832 ± 0.0003            \\\midrule
\multirow{5}{*}{\rotatebox[origin=c]{90}{AP}}      & \CheckmarkBold & \XSolidBrush   & \XSolidBrush   & \XSolidBrush       & 0.5000 ± 0.0000              & 0.5000 ± 0.0000              & 0.8972 ± 0.0037               & 0.8898 ± 0.0016              & 0.5000 ± 0.0000                & 0.5000 ± 0.0000                \\
                                                   & \CheckmarkBold & \CheckmarkBold & \XSolidBrush   & \XSolidBrush       & 0.7765 ± 0.0089              & 0.7616 ± 0.0022              & 0.9496 ± 0.0002               & 0.9355 ± 0.0003              & 0.8081 ± 0.0015                & 0.8214 ± 0.0030                \\
                                                   & \CheckmarkBold & \XSolidBrush   & \CheckmarkBold & \XSolidBrush       & 0.8501 ± 0.0036              & 0.8223 ± 0.0104              & 0.9532 ± 0.0011               & 0.9409 ± 0.0011              & 0.8862 ± 0.0157                & 0.9516 ± 0.0005                \\
                                                   & \CheckmarkBold & \XSolidBrush   & \XSolidBrush   & \CheckmarkBold     & 0.9196 ± 0.0021              & 0.9569 ± 0.0002              & 0.9677 ± 0.0003               & 0.9709 ± 0.0001              & 0.9373 ± 0.0022                & 0.9805 ± 0.0002                \\ 
                                                    & \CheckmarkBold & \CheckmarkBold   & \CheckmarkBold   & \CheckmarkBold     & 0.9527 ± 0.0008  & 0.9607 ± 0.0026  & 0.9792 ± 0.0001  & 0.9781 ± 0.0003  &    0.9483 ± 0.0021   &    0.9815 ± 0.0004             \\\bottomrule
\end{tabular}
}

\label{tab:abl_feat}

\end{table*}

\subsection{Future Link Prediction}
\label{sec:exp_complx}
In our main experiments, we compare our method with several previous SOTA models in the DTDG domain using the MRR metric, and we additionally report the AUC-ROC and AP for our experimental results. For the baseline models, where some datasets have been previously addressed, we follow the settings and results from these prior studies and those reported in ROLAND. We use historical snapshot data as training data and predict the existence of edges in future snapshots. During training, data from later snapshots could serve as a self-supervision signal for earlier snapshots. Ultimately, we select the model that performed best on the validation set for testing and predict future links on the test set.

The experimental results are shown in Table \ref{tab:baselines}. For our method, we chose three different strategies to aggregate the Neighbor Intersect Feature. We select the best experimental results for this table, and a detailed comparison will be discussed in Section \ref{sec:different_sif}. The results demonstrate that our method exceeded the performance of existing SOTA models on most datasets. On average, there is a 61.29\% improvement over previous SOTA models across the 6 datasets.
The baseline methods already perform quite well on the Reddit-Title and Reddit-Body due to the presence of edge features. Consequently, the improvement of our method over the baseline is not as substantial on these two datasets compared to others. This result can also be explained through our ablation study in Section \ref{sec:abl_feat}.
We believe that the performance improvement of our model stems from the precise modeling of node neighbor information and the intersections between the two nodes being predicted. Our method allows the final node embeddings to not only aggregate neighbor node information but also capture patterns of historical interactions. Additionally, the multi-patching module enables node representation at different granularities, allowing the model to retain local information while also capturing global-level information. The attention mechanism selectively transmits this information, enhancing the overall model's expressiveness and accuracy.

Additionally, while many previous methods frequently encountered OOM issues, our approach can easily be trained on larger datasets such as Reddit-Title and Mathoverstack. We analyze the space complexity of our model and compared with previous methods. Our model uses Transformers as the backbone, and its space complexity is given by 
$O(L \cdot (5 \cdot d_C + d_C^2 + h \cdot \lambda_m^2))$, 
where \(h\) denotes the number of self-attention heads used in the model. It is evident that the space complexity of our model is primarily dominated by \(\lambda_m\). Thanks to the introduction of the multi-patching module, this value can be controlled and is significantly smaller than the total number of nodes, \textit{i.e.,} \(|V|\).
In contrast, a commonly used GCN model in baseline methods has a space complexity of $O(d_N \cdot (|V| + |E| + L \cdot d_h + L \cdot |V|))$, 
where \(d_h\) is the dimension of the hidden layer in the RNN. This complexity is mainly dominated by the number of nodes and edges in the input graph.
For models using an RNN to model temporal attributes, the space complexity of the simplest RNN model is $O(d_h^2 + T \cdot d_h)$, 
primarily dominated by the number of snapshots, \(T\). DTDG representation learning models under the GNN+RNN framework need to store intermediate activation values at each snapshot, resulting in higher space complexity than standalone GCNs, with a complexity of $O(d_N \cdot (|V| + |E| + L \cdot d_h + T \cdot |V|))$.
Given that \(|V| \gg \lambda_m\) and \(T\) is often on the same order of magnitude or larger than \(\lambda_m\), the space complexity of GNN+RNN models is much higher than that of our model. This explains why some previous models encounter OOM issues when handling the datasets used in this study, whereas our model leaves ample GPU memory for these experiments.

\useunder{\uline}{\ul}{}
\begin{table*}[]

\centering

\caption{Ablation Study: Applying different modes, i.e., GRU mode, MLP mode and SUM mode, for modeling the Neighbor Intersect Feature and comparing the performance to the best baselines.}

\begin{tabular}{c|c|cccccc}
\toprule
Metric                                                          & Mode    & Bitcoin-OTC           & UCI-Message           & Reddit-Title        & Reddit-Body         & Mathoverstack       & Email                 \\ \midrule
MRR & Baseline & {\ul 0.2203 ± 0.0167} & {\ul 0.2289 ± 0.0618} & {\ul 0.425 ± 0.015} & {\ul 0.362 ± 0.002} & {\ul 0.271 ± 0.045} & {\ul 0.1024 ± 0.0261} \\ \midrule
\multirow{3}{*}{\rotatebox[origin=c]{90}{MRR}}                  
                                                                & GRU      & 0.2347 ± 0.0190       & 0.2276 ± 0.0066       &  0.3189 ± 0.0486          &  0.2845 ± 0.0072    &      N/A, OOM       &   0.2191 ± 0.0043      \\
                                                                & MLP      & 0.2313 ± 0.0181       & 0.3479 ± 0.0322       & 0.2929 ± 0.0033     & 0.2845 ± 0.0042     & 0.5431 ± 0.0328     & 0.2398 ± 0.0226       \\
                                                                & SUM      & 0.2838 ± 0.0291       & 0.4925 ± 0.0208       & 0.4296 ± 0.0114     & 0.5055 ± 0.0166     & 0.4864 ± 0.0279     & 0.2012 ± 0.0107       \\ \midrule
\multirow{3}{*}{\rotatebox[origin=c]{90}{\makecell{AUC-\\ROC}}} & GRU      & 0.9371 ± 0.0022       & 0.9174 ± 0.0032       &  0.9617 ± 0.0004    &   0.9525 ± 0.0016    &      N/A, OOM         &    0.9646 ± 0.0006   \\
                                                                & MLP      & 0.9368 ± 0.0044       & 0.8955 ± 0.0178       & 0.9617 ± 0.0013     & 0.9523 ± 0.0009     & 0.9745 ± 0.0018     & 0.9670 ± 0.0006       \\
                                                                & SUM      & 0.9477 ± 0.0022       & 0.9524 ± 0.0024       & 0.9764 ± 0.0001     & 0.9746 ± 0.0002     & 0.9350 ± 0.0023     & 0.9832 ± 0.0003       \\ \midrule
\multirow{3}{*}{\rotatebox[origin=c]{90}{AP}}                   & GRU      & 0.9412 ± 0.0026       & 0.9220 ± 0.0022       &   0.9640 ± 0.0007     &  0.9561 ± 0.0011     &     N/A, OOM            &   0.9600 ± 0.0009     \\
                                                                & MLP      & 0.9405 ± 0.0040       & 0.9039 ± 0.0150       & 0.9641 ± 0.0008     & 0.9556 ± 0.0008     & 0.9780 ± 0.0014     & 0.9633 ± 0.0012       \\
                                                                & SUM      & 0.9527 ± 0.0008       & 0.9607 ± 0.0026       & 0.9792 ± 0.0001     & 0.9781 ± 0.0003     & 0.9483 ± 0.0021     & 0.9815 ± 0.0004       \\ \bottomrule
\end{tabular}

\label{tab:abl_intse}

\end{table*}

\subsection{An Empirical Study for Multi-Patching}
The multi-patching module allows our model to capture information about neighboring nodes at different granularities, enabling it to extract both local and global information simultaneously. However, the impact of information at different granularities varies across datasets due to differences in data distribution and size. Therefore, we conduct an empirical study using multiple patch sizes to identify the optimal number of patch sizes for different datasets. We performed experiments on the Bitcoin-OTC, UCI-Message, and Reddit-Body datasets, using up to five different patch sizes (2, 4, 8, 16, and 32). For experiments with fewer patch sizes, we chose the largest $m$ sizes from these five. The maximum input sequence length for the Transformer encoder is extended to 256.

The experimental results for these 3 datasets are shown in Figure \ref{fig:mp}. The effects of using varying numbers of patch sizes differ across datasets. However, in general, using more patch sizes helps the model capture information at different granularities, particularly for larger datasets like Reddit-Body. We also find that setting the number of patch sizes to 3 produces generally stable results, leading us to apply 3 different patch sizes in our main experiment. The optimal number of patch sizes may vary depending on the dataset, and should be determined experimentally. % It's important to note that using more patch sizes requires additional corresponding Transformer encoders, which increases model complexity but can potentially improve performance on downstream tasks.

\balance

\subsection{Ablation Study}
\subsubsection{Features}
\label{sec:abl_feat}
We conduct ablation studies to identify the sources of performance improvements associated with the five features proposed. The multi-patching module is applied across all models to assess the individual impact of these features on performance. The results, presented in Table \ref{tab:abl_feat}, confirm that each of the five features contributes positively to the model's effectiveness. Notably, as some datasets do not include Node Features and/or Edge Features, we replace these with zero vectors. Consequently, models relying solely on these two features exhibit negligible effects for those datasets.

\subsubsection{Different Modes for Modeling Neighbor Intersect Feature.}
\label{sec:different_sif}

Our approach significantly improves upon previous methods through the development of the Neighbor Intersect Feature, designed to capture the intersections between neighboring nodes. To represent this feature, we explore various modes as the $f(\cdot)$ mentioned in Section \ref{sec:ints_feat}. Experimental results are presented in Table \ref{tab:abl_intse}.
Considering that this feature generates a sequence for each neighboring node, an intuitive solution is to utilize a sequence model as $f(\cdot)$. Accordingly, we conduct experiments using the GRU model to process the sequence (GRU mode). However, this mode increases complexity and computational demands, leading to OOM errors on the largest datasets. 
Alternatively, we evaluate simpler neural network models like the MLP for representation (MLP mode). While the performance of the MLP mode may decrease on some datasets compare to the GRU mode, its reduced computational resource requirements ensure that all datasets can be processed. 
Additionally, some datasets feature nodes with limited interactions or interactions concentrated in specific snapshots, resulting in sequences characterized by a high prevalence of zeros. This can adversely affect the representation quality. To address this, we propose another alternative mode where we aggregate the occurrences of neighboring nodes as they relate to the two nodes across all snapshots by summation, and subsequently represent this aggregated data using an MLP (SUM mode). It can be seen from Table \ref{tab:abl_intse} that the SUM mode can also be applied to large datasets and may improve performance compared to the MLP mode.
Moreover, experimental results indicate that our model consistently achieves competitive results across different representation strategies for Neighbor Intersect Feature when compared to baseline models.

\section{Limitations and Future Work}
A potential limitation of the proposed method is that, although we utilized a Transformer-based approach to avoid the OOM issues inherent in GNN+RNN based DTDG models, the inclusion of a GRU model to learn changes in neighboring nodes within each snapshot while representing the Neighbor Intersect Feature increases the model's complexity. This added complexity could still result in OOM issues for larger graphs. However, this issue can be mitigated by using simpler neural networks, such as the MLP mode or SUM mode mentioned in this paper, though this might affect model performance.
Another potential limitation is that the proposed multi-patching module also increases model complexity, which may lead to longer computation times. This issue can be addressed by reducing the number of patch sizes in the multi-patching module, but doing so could impact the model's performance on some datasets.
These limitations represent a trade-off between efficiency and effectiveness: using more complex networks can yield better results but also increases model complexity.

We believe our method leaves some open research questions in DTDG representation learning. First, our method aggregates only first-hop neighbor information as a source for node representation. Future research could consider using methods such as random walks to sample higher-order neighbors and aggregate information from a broader neighborhood. Second, our method focuses on constructing multiple features for neighboring nodes based on the properties of DTDGs. Exploring other potential features and patterns in DTDGs could be a promising research direction. Lastly, we believe recent improvements to the transformer backbone model can be directly applied to our method to improve model performance and efficiency.

\section{Conclusion}
In this paper, we introduce a novel representation learning method for DTDGs that utilizes a Transformer as the backbone model. Our approach leverages neighboring nodes to learn node representations and then conducts future link prediction as a downstream task. Innovatively, we employ snapshot or temporal information as a substitute for traditional positional encoding, and link two nodes to model their intersections within historical snapshots. This technique aims to enhance the model's ability to more effectively express node information. By capturing the dynamic intersections between nodes over time, our model provides a nuanced understanding of the network’s evolving structure. Additionally, we develop a multi-patching module designed to reduce the computational load of the model while capturing graph information at various granularities. We conduct extensive experiments on 6 public datasets, demonstrating the superior performance and scalability of our model to large datasets. Comprehensive ablation studies further validate the effectiveness of each component within our model.

\begin{acks}
This work is partially supported by the National Key Research and Development Plan Project 2022YFC3600901, the National Natural Science Foundation of China Projects No.U1936213 and NSF through grants IIS-1763365 and IIS-2106972. This work is also supported by Ant Group.
\end{acks}

\bibliographystyle{unsrt}  
\bibliography{output}

\end{document}